\documentclass[acmtog,screen,nonacm]{acmart} % arxiv
\acmSubmissionID{1225}

\usepackage{multirow,makecell}
\usepackage{mathtools}
\usepackage[capitalize,nameinlink]{cleveref}
\usepackage{xspace}
\usepackage{stfloats}
\crefname{section}{Sec.}{Secs.}
\Crefname{section}{Section}{Sections}
\Crefname{table}{Table}{Tables}
\crefname{table}{Tab.}{Tabs.}

\newcommand{\ie}{\emph{i.e.}\xspace}

\usepackage[ruled]{algorithm2e} % For algorithms

\SetAlFnt{\small}
\SetAlCapFnt{\small}
\SetAlCapNameFnt{\small}
\SetAlCapHSkip{0pt}

\acmJournal{TOG}
\newcommand{\OURS}{L3DG}

\copyrightyear{2024} % camera-ready
\acmYear{2024} % camera-ready
\setcopyright{rightsretained} % camera-ready
\acmConference[SA Conference Papers '24]{SIGGRAPH Asia 2024 Conference Papers}{December 3--6, 2024}{Tokyo, Japan} % camera-ready
\acmBooktitle{SIGGRAPH Asia 2024 Conference Papers (SA Conference Papers '24), December 3--6, 2024, Tokyo, Japan}\acmDOI{10.1145/3680528.3687699} % camera-ready
\acmISBN{979-8-4007-1131-2/24/12} % camera-ready

\begin{document}
% Title portion
\title{\OURS: Latent 3D Gaussian Diffusion}

% DO NOT ENTER AUTHOR INFORMATION FOR ANONYMOUS TECHNICAL PAPER SUBMISSIONS TO SIGGRAPH 2019!
\author{Barbara Roessle}
\affiliation{%
  \institution{Technical University of Munich}
  \country{Germany}}
\email{barbara.roessle@tum.de}
\author{Norman M{\"u}ller}
\affiliation{%
  \institution{Meta Reality Labs Zurich}
  \country{Switzerland}}
\email{normanm@meta.com}
\author{Lorenzo Porzi}
\affiliation{%
  \institution{Meta Reality Labs Zurich}
  \country{Switzerland}}
\email{porzi@meta.com}
\author{Samuel Rota Bul{\`o}}
\affiliation{%
  \institution{Meta Reality Labs Zurich}
  \country{Switzerland}}
\email{rotabulo@meta.com}
\author{Peter Kontschieder}
\affiliation{%
  \institution{Meta Reality Labs Zurich}
  \country{Switzerland}}
\email{pkontschieder@meta.com}
\author{Angela Dai}
\affiliation{%
  \institution{Technical University of Munich}
  \country{Germany}}
\email{angela.dai@tum.de}
\author{Matthias Nie{\ss}ner}
\affiliation{%
  \institution{Technical University of Munich}
  \country{Germany}}
\email{niessner@tum.de}
%\author{Gang Zhou}
%\orcid{1234-5678-9012-3456}
%\affiliation{%
%  \institution{College of William and Mary}
%  \streetaddress{104 Jamestown Rd}
%  \city{Williamsburg}
%  \state{VA}
%  \postcode{23185}
%\email{gang_zhou@wm.edu}
%\affiliation{%
%  \institution{Inria Paris-Rocquencourt}
%\email{beranger@inria.fr}
%\affiliation{%
% \institution{Rajiv Gandhi University}
%\email{aprna_patel@rguhs.ac.in}
%\affiliation{%
%  \institution{Tsinghua University}
%  \streetaddress{30 Shuangqing Rd}
%  \country{China}
%}
%\email{chan0345@tsinghua.edu.cn}
%\author{Ting Yan}
%\affiliation{%
%  \institution{Eaton Innovation Center}
%  \city{Prague}
%  \country{Czech Republic}}
%\email{yanting02@gmail.com}
%\author{Tian He}
%\affiliation{%
%  \institution{University of Virginia}
%  \department{School of Engineering}
%  \city{Charlottesville}
%  \state{VA}
%  \postcode{22903}
%  \country{USA}
%}
%\affiliation{%
%  \institution{University of Minnesota}
%\email{tinghe@uva.edu}
%\author{Chengdu Huang}
%\author{John A. Stankovic}
%\author{Tarek F. Abdelzaher}
%\affiliation{%
%  \institution{University of Virginia}
%  \department{School of Engineering}
%  \city{Charlottesville}
%  \state{VA}
%  \postcode{22903}
%  \country{USA}
%}

%\renewcommand\shortauthors{Roessle, B. et al}

\begin{abstract}
We propose \OURS, the first approach for generative 3D modeling of 3D Gaussians through a latent 3D Gaussian diffusion formulation.
This enables effective generative 3D modeling, scaling to generation of entire room-scale scenes which can be very efficiently rendered.
To enable effective synthesis of 3D Gaussians, we propose a latent diffusion formulation, operating in a compressed latent space of 3D Gaussians.
This compressed latent space is learned by a vector-quantized variational autoencoder (VQ-VAE), for which we employ a sparse convolutional architecture to efficiently operate on room-scale scenes. 
This way, the complexity of the costly generation process via diffusion is substantially reduced, allowing higher detail on object-level generation, as well as scalability to large scenes. 
By leveraging the 3D Gaussian representation, the generated scenes can be rendered from arbitrary viewpoints in real-time. 
We demonstrate that our approach significantly improves visual quality over prior work on unconditional object-level radiance field synthesis and showcase its applicability to room-scale scene generation. 
\end{abstract}





\begin{teaserfigure} % left bottom right top
  \includegraphics[width=\textwidth,trim={0.cm 20.4cm 1.3cm 0.0cm},clip]{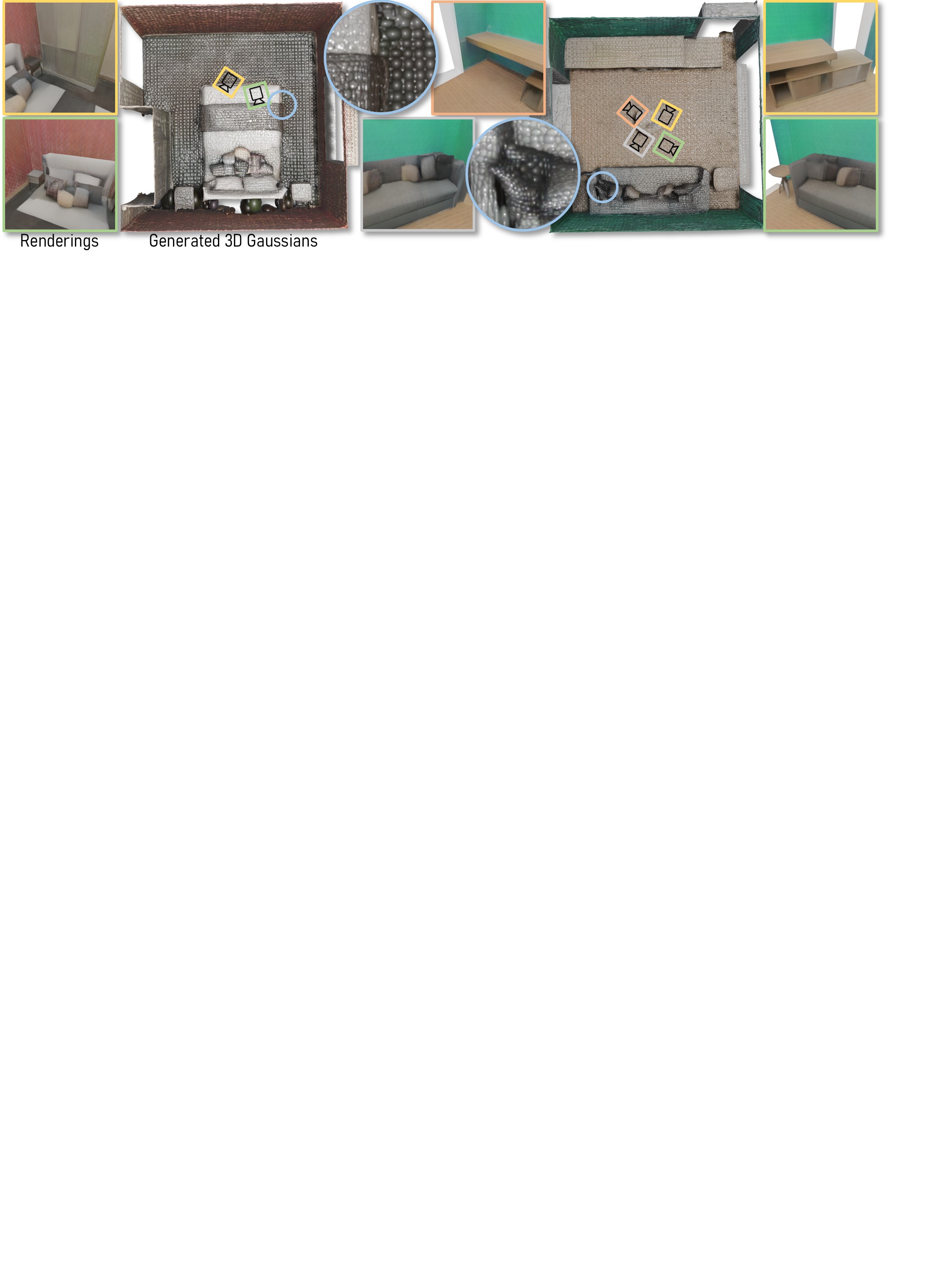}
  \caption{L3DG learns a compressed latent space of 3D Gaussian representations and efficiently synthesizes novel scenes via diffusion in latent space. This approach makes L3DG scalable to room-size scenes, which are generated from pure noise leading to geometrically realistic scenes of 3D Gaussians that can be rendered in real-time. Above results are from our model trained on 3D-FRONT; we visualize the 3D Gaussian ellipsoids and show renderings.}
  \label{fig:teaser}
\end{teaserfigure}

\maketitle

\section{Introduction}
Generation of 3D content provides the foundation for many computer graphics applications, from assest creation for video games and films to augmented and virtual reality and creating immersive visual media.
In recent years, volumetric rendering~\cite{kajiya1984ray,mildenhall2020nerf,kerbl3Dgaussians} has become a powerful scene representation for 3D content, enabling impressive photorealistic rendering, as it yields effective gradient propagation.
3D Gaussians~\cite{kerbl3Dgaussians} have become a particularly popular representation for volumetric rendering that leverages the traditional graphics pipeline in order to obtain high-fidelity renderings at real-time rates.
This combination of fast rendering speed and smooth gradients through the optimization, makes 3D Gaussians an ideal candidate for generative 3D modeling.

Inspired by the success of generative modeling for neural radiance fields of single objects \cite{muller2023diffrf,Chan2022eg3d,ssdnerf}, we aim to design a generative model for 3D Gaussians, which can provide a more scalable, rendering-efficient representation for 3D generative modeling.
Unfortunately, such generative modeling of 3D Gaussians remains challenging.
In particular, this requires a joint understanding of both scene structure as well as the intricacies of realistic appearance, for varying-sized scenes.
Moreover, 3D Gaussians are irregularly structured sets, typically containing large quantities of varying numbers of Gaussians, which a generative model must unify into an effective latent manifold.
This necessitates a flexible, scalable learned feature representation from which a generative model can be trained.

We thus propose a new generative approach for unconditional synthesis of %that synthesizes
3D Gaussians, as a representation that enables high-fidelity view synthesis for both small-scale single objects using $\sim$8k Gaussians, and enables effective scaling to room-scale scenes with $\sim$200k Gaussians.
To facilitate synthesis of large-scale environments, we formulate a latent 3D Gaussian diffusion process.
We learn a compressed latent space of 3D Gaussians on a hybrid sparse grid representation for 3D Gaussians, where each sparse voxel encodes a corresponding 3D Gaussian. %set of corresponding 3D Gaussians. 
This latent space is trained as a vector-quantized variational autoencoder (VQ-VAE), and its efficient encoding of 3D Gaussians enables flexible representation scaling from objects to 3D rooms.
We then train the generation process through diffusion on this latent 3D Gaussian space, enabling high-fidelity synthesis of 3D Gaussians representing room-scale scenes.
Experiments on both object and 3D scene data show that our approach not only produces higher quality synthesis of objects than state of the art, but also much more effectively scales to large scenes, producing 3D scene generation with realistic view synthesis. Our latent 3D Gaussian diffusion improves the FID metric by $\sim$45\% compared to DiffRF on PhotoShape ~\cite{photoshape2018}. 

In summary, our contributions are:
\begin{itemize}
\item the first approach to model 3D Gaussians as a generative latent diffusion model, enabling effective synthesis of 3D Gaussian representations of room-scale scenes that yield realistic view synthesis. 
\item our latent 3D Gaussian diffusion formulation enables flexible generative modeling on a compressed latent space constructed by sparse 3D convolutions, capturing both high-fidelity objects as well as larger, room-scale scenes.
\end{itemize}
\section{Related Work}
\begin{figure*}[tb] % left bottom right top
  \centering
\includegraphics[width=0.97\textwidth,trim={0.4cm 2.8cm 0.5cm 1.2cm},clip]{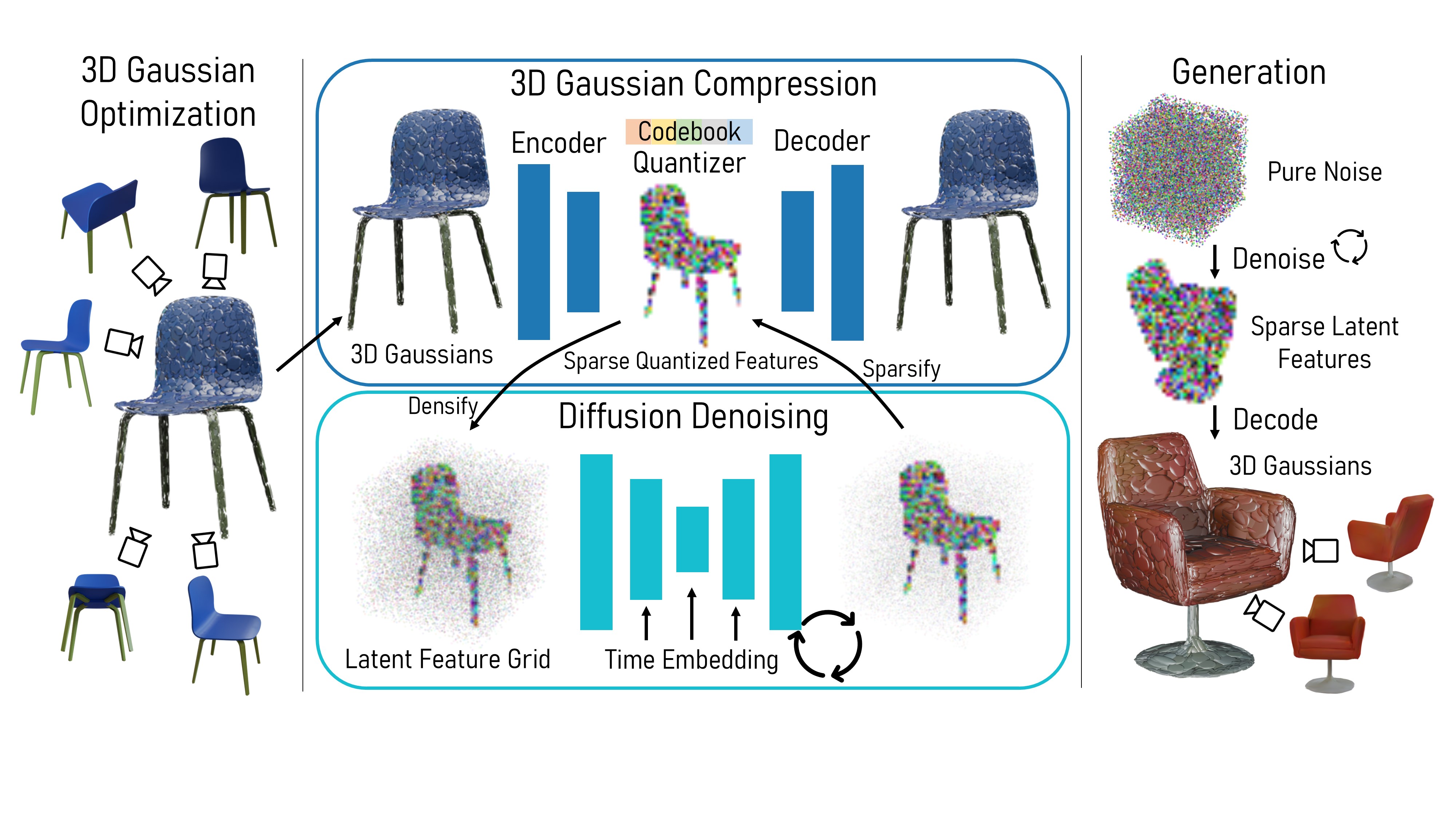}
   \caption{L3DG method overview: our 3D Gausssian compression model learns to compress 3D Gaussians into sparse quantized features using sparse convolutions and vector-quantization at the bottleneck (VQ-VAE). This allows our 3D diffusion model to efficiently operate on the compressed latent space. At test time, novel scenes are generated by denoising in latent space, which can be sparsified and decoded to high quality 3D Gaussians.}
   \label{fig:pipeline}
\end{figure*}
Our work addresses the problem of unconditional generation of 3D objects and scenes. We review below related works categorizing them based on the type of generative model.

\paragraph{GAN-based.} Generative Adversarial Networks~\cite{goodfellow2014generative} (GAN) have found successful application in the generation of 3D assets. The generator maps random noise to the target 3D representation, from which images are rendered given camera poses and supervised with the discriminator. Accordingly, methods in this category can be trained without having explicit access to 3D ground-truth, but rely only on posed images. Several 3D representations have been considered in different works ranging from simple sets of 3D primitives~\cite{liao2020unsupervised}, 3D meshes~\cite{gao2022get3d} and voxel grids~\cite{nguyenphuoc2019hologan} to radiance fields~\cite{chanmonteiro2020pi-GAN,schwarz2021graf,schwarz2022voxgraf,skorokhodov2022epigraf} and more recent Gaussian primitives~\cite{barthel2024gaussian}. Some works generate a latent 3D representation that is rendered into a 2D feature map and then decoded into the final image~\cite{Niemeyer2020GIRAFFE,gu2021stylenerf,wewer2024latentsplat}. This yields higher quality images, but at the cost of 3D inconsistencies across views.

\paragraph{Diffusion-based.} Methods in this category are built upon denoising diffusion probabilistic models~\cite{ho2020denoising} to generate 3D assets. Akin to GAN-based models, works in this category include solutions that operate on different 3D representations, requiring direct 3D observations or indirect ones (e.g., 2D images). Methods that extract 3D information from 2D images in a pre-processing step before learning the diffusion model are sometimes referred to as two-stage approaches.
Moreover, the diffusion model is either defined directly on the space of the target 3D representation, or inspired by Latent Diffusion Models~\cite{latent_diffusion} on a latent space 
that is mapped to/from the 3D representation via learned decoder/encoder pairs.
Among works that learn a 3D diffusion model from 3D observations (or have two stages), we find~\cite{cai2020learning,luo2021diffusion} operating on 3D point clouds, \cite{muller2023diffrf} operating on grid-based radiance fields, and~\cite{zhang2024gaussiancube} operating on 3D Gaussian primitives. 
Notably, \cite{ssdnerf} proposes a single-stage method that operates on the target 3D representation, namely tri-plane NeRF, but only requires indirect observations.
Examples of methods operating on a latent space include \cite{zeng2022lion,li20233dqd} for 3D point clouds and \cite{bautista2022gaudi,ntavelis2023autodecoding} for grid-based radiance fields. Similar to our work, \cite{li20233dqd} leverages a VQ-VAE~\cite{vqvae} to construct the latent space, however, their focus is on the generation of object geometries as opposed to our latent 3D Gaussian diffusion enabling object and room-level view synthesis. 
There also exists a stream of works that use diffusion models directly in image space. Among those we have methods that optimize the 3D representation given the 2D supervision generated by a text-to-image diffusion model, like~\cite{poole2022dreamfusion} for NeRFs and~\cite{li2023gaussiandiffusion,yi2024gaussiandreamer,chen2024textto3d} for Gaussian primitives. However, these methods are limited to single object generation and their per-shape optimization approach is slower than our generations in a diffusion denoising process. 
We then find methods like~\cite{anciukevičius2024renderdiffusion} that denoise images by mapping them to the 3D representation and using rendering to map back to image space. In addition, there are works that use diffusion to generate 2D views of a hypothetical 3D scene directly~\cite{watson2022novel,liu2024syncdreamer}. These latter models can produce high-quality images, but they are potentially 3D inconsistent, and require typically some form of image conditioning, although unconditional generation could be achieved by pairing it with an unconditional image generator.

Our method falls into the category of diffusion-based models that operate in latent-space with Gaussian primitives as our underlying 3D representation. Following~\cite{latent_diffusion}, our model consists of a VQ-VAE that is trained on direct 3D observations to map to/from a latent representation and a diffusion model operating on the latter space. To our knowledge, we are the first method of this kind.

\section{Method}
We focus on the task of unconditional synthesis of 3D Gaussians primitives as a high-fidelity scene representation that features real-time rendering. To enable detailed 3D generation of objects and scalability to room-size scenes, our method lifts the 3D representation of Gaussian primitives to a learned, compressed latent space on which a diffusion model can efficiently operate. The generated latent representation is learned in a feature grid that can be decoded back to a set of 3D Gaussian primitives to support fast novel-view synthesis (\cref{fig:pipeline}). To efficiently map between the Gaussian primitives (\cref{ssec:3dgaussians_preliminaries}) and the latent representation on which the diffusion model operates, we introduce a sparse convolutional network, which implements a VQ-VAE~(\cref{ssec:vqvae}). Finally, our latent diffusion model learns a denoising process in our low-dimensional latent space to unconditionally generate novel 3D Gaussian scenes from pure noise (\cref{ssec:latent_3dgaussian_diffusion}).

\subsection{Preliminaries: 3D Gaussian Splatting}
\label{ssec:3dgaussians_preliminaries}
Given a set of RGB images with camera poses, 3D Gaussian Splatting (3DG)~\cite{kerbl3Dgaussians} reconstructs the corresponding static scene, represented as a collection of 3D Gaussian primitives. 
Each Gaussian primitive comprises a 3D position $\boldsymbol{\mu}_i\in\mathbb R^3$ and a 3D covariance matrix $\Sigma_i$ that is factorized as $\Sigma_i\coloneqq R_iS_i^2R_i^\top$,
where $S_i$ is a nonnegative, diagonal scale matrix with diagonal denoted by $\boldsymbol s_i\in\mathbb R_+^3$, and $R_i\in \text{SO}(3)$ is a rotation matrix represented as a unit quaternion $\mathbf{r}_i \in \mathbb{R}^4$.
To support rendering of RGB images, the view-dependent color $\mathbf{c}_i\left(\boldsymbol{\gamma}_i, \mathbf{d}\right)\in\mathbb{R}^3$ of each Gaussian primitive is obtained from its spherical harmonics coefficients $\boldsymbol{\gamma}_i$ and the viewing direction $\mathbf{d}$. 
In addition, each Gaussian primitive entails an opacity $\alpha_i \in \mathbb{R}_+$. 
An image $C_\pi$ from camera $\pi$ can be rendered by projecting and blending $N$ depth-ordered Gaussians primitives as follows:
\begin{equation}
    C_\pi(\mathbf u) \coloneqq \sum_{i=1}^N\mathbf{c}_i\left(\boldsymbol{\gamma}_i, \mathbf{d}\right)\, \omega_\pi^i(\mathbf u) \prod_{j=1}^{i-1}[1-\omega_\pi^j(\mathbf u)],
\end{equation}
where  $\omega^i_\pi(\mathbf u)\coloneqq \alpha_i G_\pi^i(\mathbf u)$ is the opacity of the $i$th primitive scaled by the contribution of the following function:
\begin{equation}
G_\pi^i(\mathbf u)\coloneqq\exp\left[-\frac{1}{2}(\mathbf u-\boldsymbol\mu^\pi_i)^\top (\Sigma^\pi_i)^{-1}(\mathbf u-\boldsymbol\mu^\pi_i)\right]\,.    
\end{equation}
This represents the kernel of the 2D Gaussian with parameters $(\boldsymbol{\mu}_i^\pi,\Sigma_i^\pi)$ that we obtain when projecting the primitive's 3D Gaussian with parameters $(\boldsymbol\mu_i,\Sigma_i)$ to the camera image plane under a linear approximation of the projection function (see~\cite{kerbl3Dgaussians} for more details).

The parameters of the 3D Gaussian primitives of a scene are optimized by minimizing an $L_1$ color loss $\mathcal{L}_\mathrm{RGB}$ and the negated structural similarity metric (SSIM)~\cite{wang2004image} between rendered images $\hat{I}$ and target images $I$: 
\begin{equation}
    \mathcal{L}_{\mathrm{3DG}} \coloneqq (1 - \lambda_{\mathrm{3DG}}) \mathcal{L}_{\mathrm{RGB}} + \lambda_{\mathrm{3DG}} (1 - \text{SSIM}(\hat{I}, I)),
    \label{eq:l3dg}
\end{equation}
where $\lambda_{\mathrm{3DG}}$ is a balancing factor. 
\subsection{Learning a Latent Space for 3D Gaussians}
\label{ssec:vqvae}
While 3D Gaussian Splatting offers an explicit representation that is highly expressive and efficient, the point cloud of 3D Gaussian primitives is spatially unstructured and sparse.
The unstructured nature makes it challenging for a generalized model to learn. 
To recover spatial structure, we optimize primitives that are assigned to voxels of a sparse grid 
(\cref{sssec:grid_aligned_3dg}). To cope with sparsity, we define a network comprising sparse convolutions to compress our 3D representation (\cref{sssec:sparse_vqvae}) into a latent dense grid of low spatial resolution. 
\subsubsection{Sparse Grid-assigned 3D Gaussians}
\label{sssec:grid_aligned_3dg}
To train our VQ-VAE, we pre-compute for each scene 3D Gaussian primitives that are aligned with a sparse grid, i.e, the space of a scene is discretized into a 3D grid with voxel size $d$ and each primitive is \emph{uniquely} assigned to a voxel. This representation is optimized similar to~\cite{kerbl3Dgaussians} with a few differences.
First, the position of a primitive $\boldsymbol{\mu}_i$ is reparametrized in terms of a voxel index $\kappa_i$ and a 3D displacement $\psi(\boldsymbol\delta_i)$, depending on $\boldsymbol\delta_i\in\mathbb R^3$, so that the primitive's center becomes $\boldsymbol\mu_i\coloneqq\mathbf y_{\kappa_i}+\psi(\boldsymbol\delta_i)$, where $\mathbf y_j\in\mathbb R^3$ denotes the 3D center of the $j$th voxel. Since each voxel can be assigned at most one Gaussian primitive, the set of Gaussian primitives can be represented as a sparse grid $\boldsymbol\theta$ of Gaussian parameters, where the $\kappa_i$th cell denoted by $\boldsymbol\theta_{\kappa_i}$ contains the parameters of the $i$th Gaussian primitive, \ie $\boldsymbol\theta_{\kappa_i}\coloneqq (\boldsymbol\delta_i,\mathbf s_i,\mathbf r_i,\boldsymbol\gamma_i,\alpha_i)$.
During optimization, 3D primitives can move within their voxel and adjacent ones. To enforce this, we set $\psi(\boldsymbol{\delta}) \coloneqq 1.5\,\text{tanh}(\boldsymbol{\delta})\,d$. 
Second, we introduce a new densification strategy. A new Gaussian primitive is created in an inactive voxel, if an existing primitive from a neighboring cell moves into that voxel and the magnitude of its averaged view-space positional gradient exceeds a threshold $\epsilon_{\delta}$, indicating the need for densification. The newly-created primitive is initialized at the center of the voxel (zero displacement) with isotropic scale $d$, identity rotation matrix, predefined small opacity and appearance averaged from all primitives competing for densification on the same voxel.

Akin to~\cite{kerbl3Dgaussians}, primitives with opacity below a threshold $\epsilon_{\alpha}$ are pruned and the proposed sparse, grid-assigned representation of 3D Gaussian primitives is optimized by minimizing $\mathcal{L}_{\mathrm{3DG}}$ as defined in \cref{eq:l3dg}.
\subsubsection{3D Gaussian Compression Model}
\label{sssec:sparse_vqvae}
Our 3D Gaussian compression model is inspired by the success of latent diffusion~\cite{latent_diffusion} for image synthesis. Specifically, we employ a VQ-VAE~\cite{vqvae} due to its ability to learn an expressive prior over a small, discretized latent space, which is particularly valuable to handle the complexity of 3D space. Our network leverages 3D sparse convolutions to map between the sparse, grid-assigned Gaussians and a small latent dense grid. A vector quantization layer with a codebook of size $K$ is employed at the bottleneck between encoder $E$ and decoder $D$: 
\begin{align}
    \mathbf{z}_e &\coloneqq E(\boldsymbol{\theta}), \\
    \mathbf{z}_q &\coloneqq \text{quantize}(\mathbf{z}_e), \\
    \hat{\boldsymbol{\theta}} &\coloneqq D(\mathbf{z}_q),
\end{align}
where $\mathbf{z}_e$ is the output of the encoder given the sparse, grid-assigned 3D Gaussians $\boldsymbol\theta$, $\mathbf{z}_q$ is the quantized sparse latent space and $\hat{\boldsymbol{\theta}}$ is the reconstructed representation. $\mathbf{z}_q$ serves as input to the diffusion model (\cref{ssec:latent_3dgaussian_diffusion}), where it is converted to a low resolution dense grid. The 3D Gaussian compression network is trained with a VQ-VAE commitment loss $\mathcal{L}_{\mathrm{commit}}$, to ensure the encoder commits to embeddings in the codebook. As reconstruction losses, we employ an $L_1$ color loss $\mathcal{L}_\mathrm{RGB}$ and a perceptual loss $\mathcal{L}_{\mathrm{perc}}$ on $M$ renderings of the reconstructed 3D Gaussians from different viewpoints. The perceptual loss encourages similarity of reconstructed and target features at different levels of detail. 

The decoder employs generative sparse transpose convolutions \cite{gwak2020gsdn} to enable the generation of new coordinates in the upsampling. This is crucial to allow standalone usage of the decoder to decode synthesized latent grids from the diffusion model, without the possibility of leveraging cached coordinates from the encoder as in standard sparse transpose convolutions. The generation of new coordinates in each upsampling layer comes with the need for a pruning strategy to avoid an explosion in the number of active voxels. Thus, after each upsampling, a linear layer classifies each predicted voxel as occupied or free~\cite{ogn2017}. During training, these occupancies are supervised with a binary cross entropy loss (BCE) $\mathcal{L}_{\mathrm{occ}}$ using the grid-assigned 3D Gaussians as target. At test time, they serve to effectively prune the predicted voxels. Thus, we define the combined loss function of the 3D Gaussian compression model as follows:
\begin{align}
    \mathcal{L}_{\mathrm{comp}} \coloneqq \lambda_{\mathrm{commit}}&\mathcal{L}_{\mathrm{commit}} + \lambda_{\mathrm{RGB}}\mathcal{L}_{\mathrm{RGB}} + \lambda_{\mathrm{perc}}\mathcal{L}_{\mathrm{perc}} + \mathcal{L}_{\mathrm{occ}}, \\
    \text{where} \quad &\mathcal{L}_{\mathrm{commit}} \coloneqq \Vert \mathbf{z}_e - \mathbf{e}_{\bot}\Vert_2^2, \\
    &\mathcal{L}_{\mathrm{perc}} \coloneqq \Vert \Phi_\mathrm{VGG}(\hat{I})-\Phi_\mathrm{VGG}(I)\Vert_2^2\,,
\end{align}
where $\mathbf{e}$ are codebook entries and we use $\bot$ to indicate that gradients to the embeddings are stopped. The codebook items are instead updated using an exponential moving average~\cite{vqvae}. $\Phi_\mathrm{VGG}$ is the vectorized concatenation of the first 5 feature layers before the max pooling operation of a VGG19 network~\cite{vgg}, where each layer is normalized by the square root of the number of elements. 

The 3D Gaussian compression model learns a compact representation, where two downsampling layers of stride 2 lead to a volumetric compression by a factor of $64$. 
At the same time, the number of parameters per voxel is drastically reduced to 4-element codebook items, where the codebook size is kept below 10k in all experiments. 
\subsection{Latent 3D Gaussian Diffusion}
\label{ssec:latent_3dgaussian_diffusion}
We propose a latent 3D diffusion model to learn the distribution of the compact latent space of 3D Gaussians $p$. 
To generate scenes from pure noise, without any prior knowledge, we need to use a dense grid in the latent diffusion, such that content may be synthesized anywhere in space. Hence, the compressed sparse grid is first converted to the corresponding low-resolution dense grid. To enable switching back to the sparse representation for decoding, the diffusion model is trained to denoise an additional occupancy element in the dense form. 

The generation process is an inverse discrete-time Markov forward process. The forward process repeatedly adds Gaussian noise $\boldsymbol{\epsilon} \in \mathcal{N}(\mathbf{0}, \mathbf{I})$ to a sample $\mathbf{z}_0 \sim p$ leading to a series of increasingly noisy samples $\left\{\mathbf{z}_t | t \in [0,T]\right\}$. The noisy sample $\mathbf{z}_t$ at a time step $t$ is defined as 
\begin{equation}
    \mathbf{z}_t \coloneqq \alpha_t \mathbf{z}_0 + \sigma_t \boldsymbol{\epsilon},
\end{equation}
where parameters $\alpha_t$ and $\sigma_t$ determine the amount of noise as part of the noise scheduling. After $T$ noising steps the sample becomes pure Gaussian noise (i.e., $\alpha_T\approx 0$ and $\sigma_T\approx 1$). Our diffusion model reverses the forward process, i.e., it iteratively denoises a noisy sample beginning at $T$ with pure noise, yielding at the end a clean sample $\mathbf{z}_0$. 
We train our diffusion model $\hat{\mathbf{v}}_\phi(\mathbf{z}_t, t)$, parametrized by $\phi$, to perform $\mathbf{v}$-prediction~\cite{salimans2022progressive}, where the network output relates to the predicted clean sample $\hat{\mathbf{z}}_0$ by 
\begin{equation}
    \hat{\mathbf{z}}_0 \coloneqq \alpha_t \mathbf{z}_t - \sigma_t \hat{\mathbf{v}}_\phi(\mathbf{z}_t, t).
\end{equation}
This property is used to compute a mean squared error (MSE) loss against the noise-free sample to supervise the diffusion model:
\begin{equation}
    \mathcal{L}_{\mathrm{diff}} = \Vert \hat{\mathbf{z}}_0 - \mathbf{z}_0 \Vert_2^2.
\end{equation}
With our latent 3D Gaussian diffusion model the iterative synthesis process efficiently takes place in the low resolution latent space. A generated latent sample is sparsified using the occupancy channel, and decoded to a high fidelity sparse 3D Gaussian representation using the decoder from \cref{sssec:sparse_vqvae}.

\subsection{Implementation Details}
\paragraph{Sparse Grid-assigned 3D Gaussians}
On all datasets, we scale the point cloud that is used to initialize the 3D Gaussians optimization into a unit cube and use a voxel size $d=0.008$. While they use the same sparse grid resolution, scenes typically require $\sim$200k Gaussians whereas $\sim$8k are sufficient for objects. 
We set the densification and pruning thresholds $\epsilon_{\delta}=0.0008; \epsilon_{\alpha}=0.005$ and use $\lambda_{3DG}=0.2$. View dependence is modeled with spherical harmonics of degree 1, which performs best across datasets. The optimization of 3D Gaussians requires $\sim$4min/shape and $\sim$10min/room. Note that our implementation is unoptimized and significant improvement could be made through parallel processing. 

\paragraph{3D Gaussian Compression Model}
We use Minkowski Engine~\cite{choy2019minkowski} to implement the 3D sparse convolutional network. Our convolutional blocks all use kernel size 3, with batch norm and ReLU activations. The encoder starts with a convolutional block increasing the input channels to 128. It is followed by two downsampling blocks, each consisting of two residual blocks, where the first doubles the number of channels, followed by a convolutional block with stride 2. Another residual block is employed in the bottleneck, where the number of channels is 512. A convolutional layer reduces the number of channels to 4 in the latent space, where the vector quantization is applied using codebook size $K=4096$ on objects and $K=8192$ on rooms. The decoder starts with a convolutional block that increases the number of channels to 512. This is followed by 2 upsampling blocks, each consisting of two residual blocks, where the first halves the number of channels, followed by a generative transpose convolution block~\cite{gwak2020gsdn} with stride 2. After the upsampling, 2 residual blocks and a final convolutional layer map from 128 channels to the number of Gaussian parameters. We use loss weighting $\lambda_{\mathrm{commit}}=0.25$ on all datasets, $\lambda_{\mathrm{RGB}}=12.5; \lambda_{\mathrm{perc}}=0.1$ with $M=4$ images on objects and $\lambda_{\mathrm{RGB}}=7.5; \lambda_{\mathrm{perc}}=0.3$ with $M=12$ images on rooms. The batch size is 16 on objects and 4 on rooms. 
We train the model for 130/200/100 epochs on PhotoShape/ABO/3D-FRONT, using the Adam optimizer~\cite{Kingma2015AdamAM} with learning rates 0.0001/0.0002/0.0001 which are exponentially decayed by a factor of 0.998/0.98/0.95 at the end of each epoch. The training time on a single NVIDIA A100 GPU is $\sim$5d/1d/3.5d on PhotoShape/ABO/3D-FRONT. 

\paragraph{3D Diffusion Model}
The diffusion model is a 3D UNet, which adapts the architecture of~\cite{dhariwal2021diffusion} to 3D. We use attention at resolutions 8 and 4 with 64 channels per head.  The diffusion model operates on a $32^3$ grid using a linear beta scheduling from 0.0001 to 0.02 in 1000 timesteps. 
We train the model for 1500/2500/3000 epochs on PhotoShape/ABO/3D-FRONT, using the Adam optimizer~\cite{Kingma2015AdamAM} with learning rate 0.0001 which is exponentially decayed by a factor of 0.998/0.9988/0.9988 at the end of each epoch. The training time on 2/1/1 NVIDIA A100 is $\sim$3.5d/1.5d/5d using batch size 64/32/16 on PhotoShape/ABO/3D-FRONT. 
For generation, we use DDPM sampling with 1000 steps. 

\section{Experiments}
In this section, we evaluate the performance of our method on unconditional generation of 3D assets. We also provide qualitative examples showcasing the ability of our method to generate room-scale scenes.
\paragraph{Datasets.}
We consider two benchmark datasets for the quantitative analysis, namely PhotoShape Chairs~\cite{photoshape2018} and Amazon Berkeley Objects (ABO) Tables~\cite{collins2022abo}\footnote{All objects for PhotoShape Chairs and ABO Tables were originally sourced from 3D Warehouse.}.
Following~\cite{muller2023diffrf}, for PhotoShape Chairs, we consider 15,576 chairs rendered from 200 views along an Archimedean spiral. 
For ABO Tables, we use the provided 91 renderings from the upper hemisphere, considering 2-3 different environment map settings per object, resulting in 1676 tables split into 1520/156 for train/test. For room-scale scene generation, we train our model on $\sim$2000 bedroom and living room style scenes from the 3D-FRONT~\cite{fu20213dfront} dataset. We render $\sim$100-500 images for training and $\sim$20-100 for testing, depending on the scene size. For all datasets, we use $512 \times 512$ images. 

\paragraph{Metrics.}
We evaluate the quality of the generated 3D assets by measuring both the quality of the rendered images and their geometric plausibility. To evaluate the quality of the renderings, we use the Frechet Inception Distance~\cite{heusel2018gans} (FID) and Kernel Inception Distance~\cite{bińkowski2021demystifying} (KID) as implemented in~\cite{obukhov2020torchfidelity}. All metrics are evaluated at $128 \times 128$ resolution. To evaluate the geometric plausibility, following~\cite{achlioptas2018learning}, we compute the Coverage Score (COV) and Minimum Matching Distance (MMD) using Chamfer Distance (CD), where the Coverage Score
measures the diversity of the generated samples, while MMD assesses the quality of the generated samples. The geometry is extracted by voxelizing the 3D Gausssians and extracting a mesh using marching cubes, so that points on the surface can be sampled. 

\paragraph{Baselines.}
We compare our method against state-of-the-art competitors that support unconditional generation of 3D assets and fall into both GAN-based and diffusion-based categories.
Among GAN-based approaches, we consider $\pi$-GAN~\cite{chanmonteiro2020pi-GAN} and EG3D~\cite{Chan2022eg3d}. 
We also compare with the diffusion-based DiffRF~\cite{muller2023diffrf}.
All methods evaluated, including ours, use the same set of rendered images for training. GAN-based methods are
trained directly on the rendered images, while DiffRF also uses per-shape radiance fields, which are pre-computed using the available posed training images.
Similarly, our method uses pre-computed sparse grid-assigned 3D Gaussians (as per \Cref{sssec:grid_aligned_3dg}).

\subsection{Baseline Comparison}
\begin{figure*}[tbp] % left bottom right top
 \setlength\tabcolsep{0pt}
  \centering
  \begin{tabular}{>{\centering\arraybackslash}p{0.3333\textwidth}>{\centering\arraybackslash}p{0.3333\textwidth}>{\centering\arraybackslash}p{0.3333\textwidth}}
\multicolumn{3}{c}{\includegraphics[width=\textwidth,trim={0.cm 0cm 0.cm 0cm},clip]{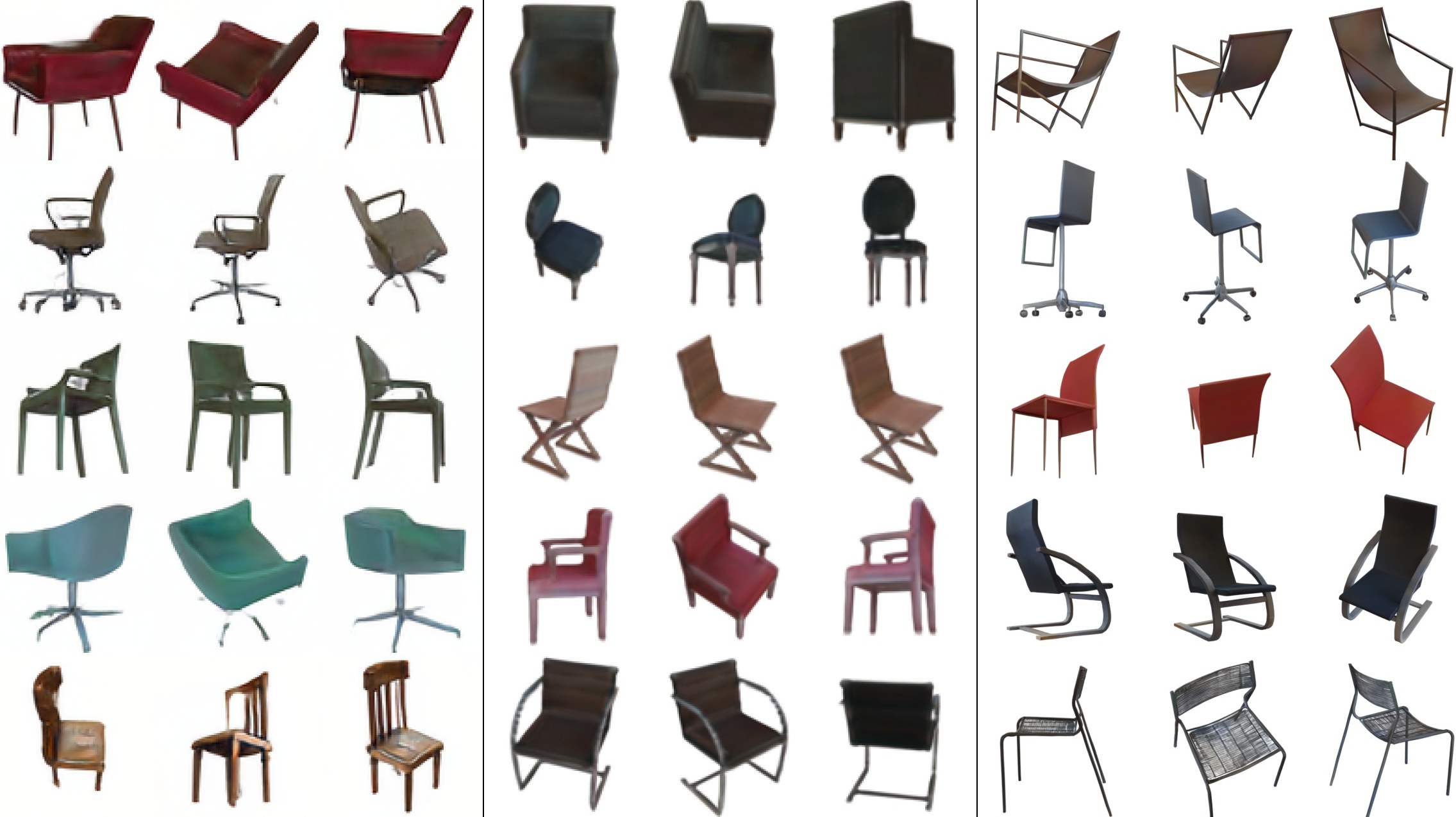}} \\
EG3D & DiffRF & Ours \\
  \end{tabular}
   \caption{Comparison on PhotoShape \cite{photoshape2018}. Our method generates more detail than the baselines, such as thin structures, and has fewer artifacts. 
   }
   \label{fig:photoshape_baseline}
\end{figure*}
\begin{table}[tb]
\centering
\captionof{table}{Quantitative comparison of unconditional generation on the PhotoShape Chairs~\cite{photoshape2018} dataset. MMD and KID scores are multiplied by $10^3$.}
\resizebox{\columnwidth}{!}{
\begin{tabular}{lccccc}
\toprule
Method  & FID $\downarrow$ & KID $\downarrow$  & COV $\uparrow$  & MMD $\downarrow$ \\ 
\midrule
$\pi$-GAN~\cite{chanmonteiro2020pi-GAN} & 52.71 & 13.64 & 39.92 & 7.387 \\
EG3D~\cite{Chan2022eg3d}                & 16.54 & 8.412 & 47.55 & 5.619 \\ 
DiffRF~\cite{muller2023diffrf}          & 15.95 & 7.935 & 58.93 & 4.416 \\
Ours                                    & \textbf{8.49} & \textbf{3.147} & \textbf{63.80} & \textbf{4.241} \\
\bottomrule
\end{tabular}
}
\label{tab:photoshape}
\end{table}
\begin{table}[tb]
\centering
\captionof{table}{Quantitative comparison of unconditional generation on the ABO Tables~\cite{collins2022abo} dataset. MMD and KID scores are multiplied by $10^3$.}
\resizebox{\columnwidth}{!}{
\begin{tabular}{lccccc}
\toprule
Method  & FID $\downarrow$ & KID $\downarrow$  & COV $\uparrow$  & MMD $\downarrow$ \\ 
\midrule
$\pi$-GAN~\cite{chanmonteiro2020pi-GAN} & 41.67   & 13.81 & 44.23  & 10.92 \\
EG3D~\cite{Chan2022eg3d}                & 31.18  & 11.67 &  48.15  &  9.327 \\ 
DiffRF~\cite{muller2023diffrf}          & 27.06 & 10.03   & 61.54  & 7.610 \\
\midrule
Ours                                    & \textbf{14.03} & \textbf{3.15} & \textbf{65.38} & \textbf{7.312} \\
\;\;w/o compression model              & 197.1 & 166.8 & 51.92 & 8.483 \\
\;\;w/o RGB loss                       & 17.26 & 3.28  & 63.46 & 7.756 \\
\;\;w/o perceptual loss                    & 34.35 & 13.65 & 61.53 & 7.488 \\
\bottomrule
\end{tabular}
}
\label{tab:abo}
\end{table}
\begin{figure*}[p] % left bottom right top
\setlength\tabcolsep{0pt}
\centering
  \begin{tabular}{>{\centering\arraybackslash}p{0.3333\textwidth}>{\centering\arraybackslash}p{0.3333\textwidth}>{\centering\arraybackslash}p{0.3333\textwidth}}
\multicolumn{3}{c}{\includegraphics[width=\textwidth,trim={0.cm 0.cm 0.0cm 0.0cm},clip]{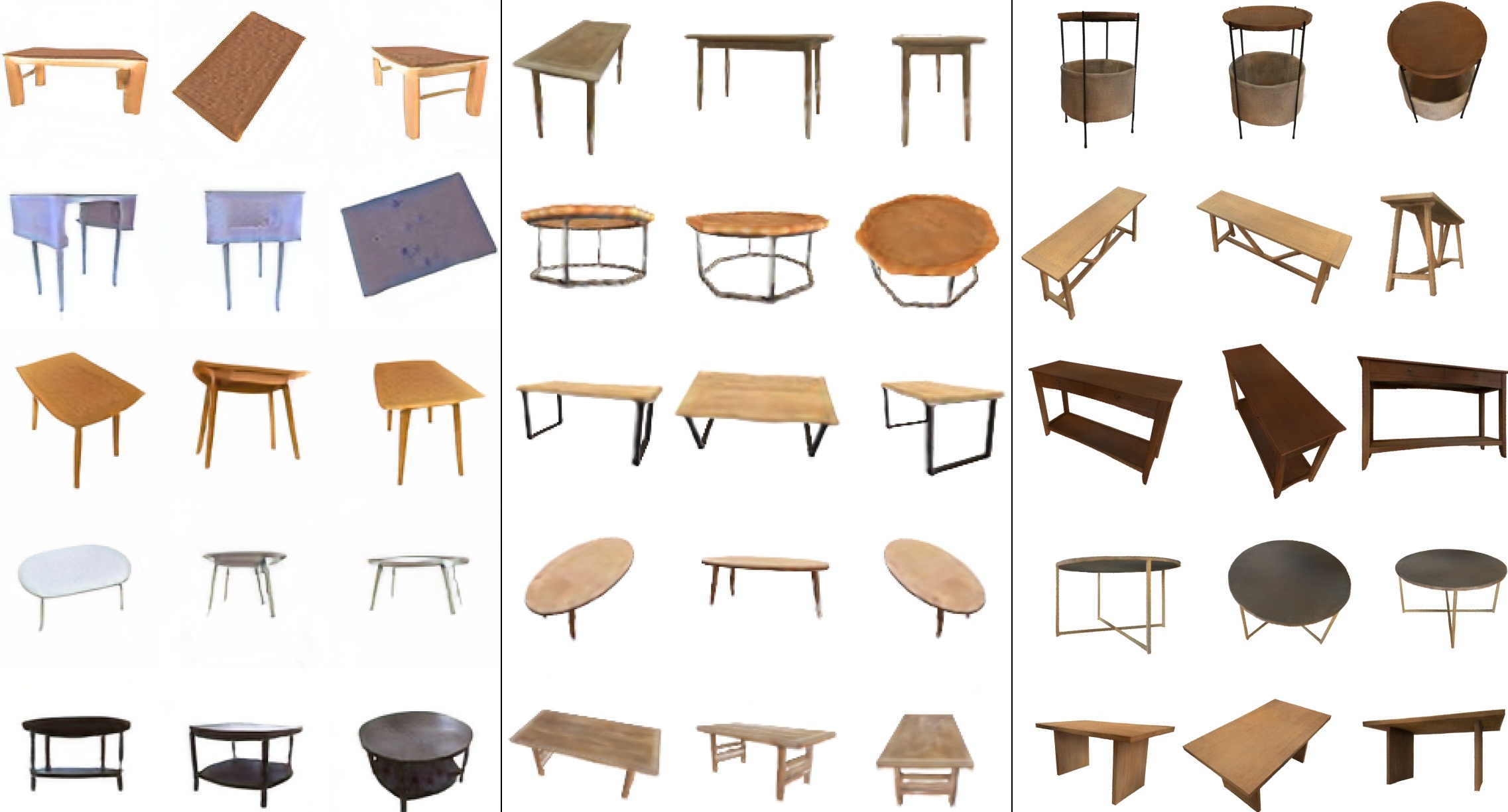}} \\[-0.1cm]
EG3D & DiffRF & Ours \\
  \end{tabular} 
  \vspace{-0.3cm}
   \caption{Comparison on ABO \cite{collins2022abo}. Tables generated by our method are sharper and show less artifacts compared to the baselines. 
   }
   \label{fig:abo_baseline}
\end{figure*}
\begin{figure*}[p]
\setlength\tabcolsep{0pt}
\centering
  \begin{tabular}{>{\centering\arraybackslash}p{0.3333\textwidth}>{\centering\arraybackslash}p{0.3333\textwidth}>{\centering\arraybackslash}p{0.3333\textwidth}}
\multicolumn{3}{c}{\includegraphics[width=\textwidth,trim={0.cm 0.cm 0.cm 0.cm},clip]{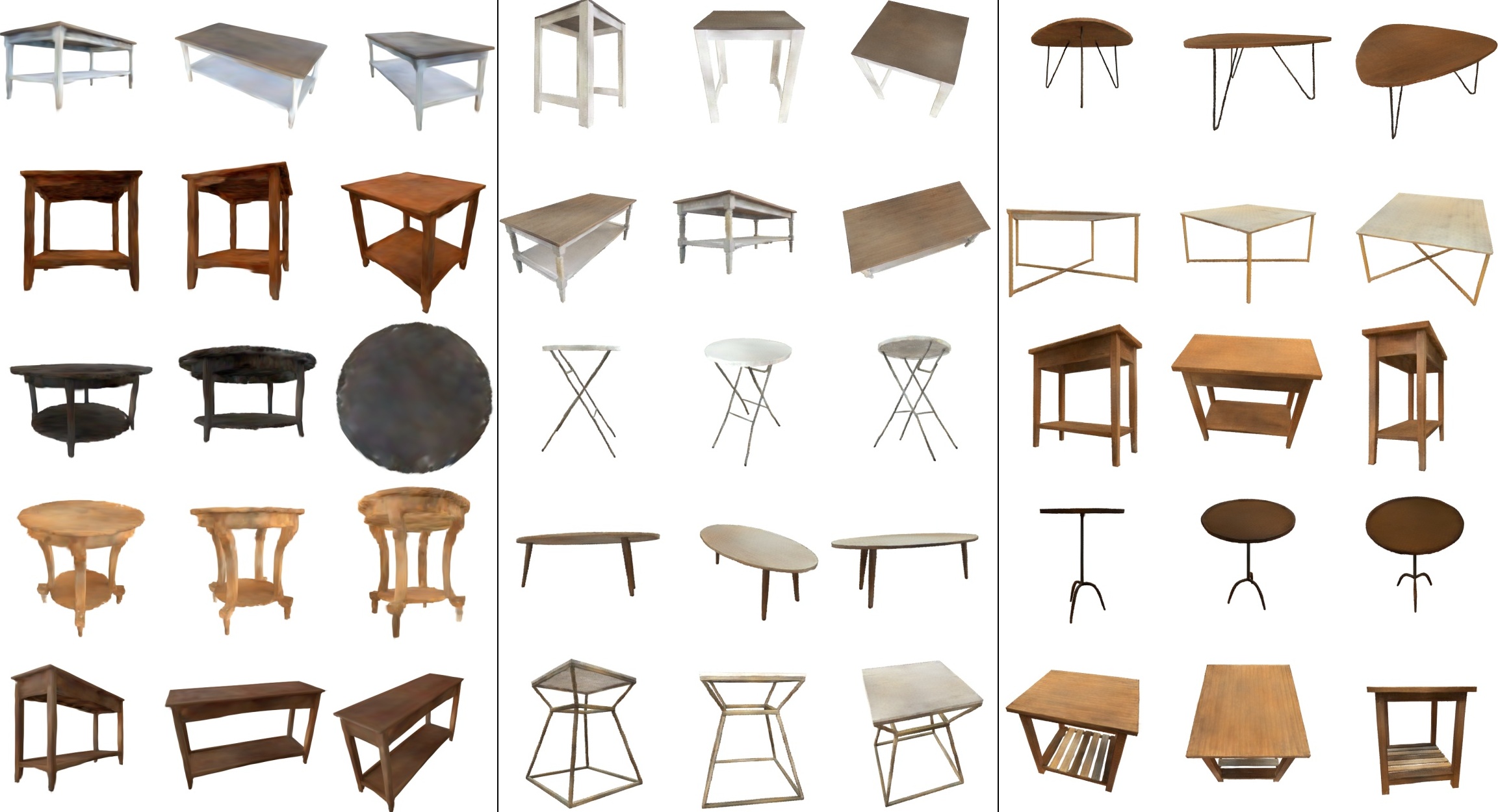}} \\[-0.1cm]
w/o perceptual loss & w/o RGB loss & Ours \\
  \end{tabular} 
  \vspace{-0.3cm}
   \caption{Ablation study on ABO \cite{collins2022abo}. Training the 3D Gaussian compression model without the rendering losses (perceptual or RGB) leads to more blurry results, especially without perceptual loss. The variant without RGB loss additionally produces less color variations in the generated scenes. 
   }
   \label{fig:abo_ablation}
\vspace{-0.3cm}
\end{figure*}
\begin{figure*}[bp] % left bottom right top
  \centering
  \includegraphics[width=0.97\textwidth,trim={1.2cm 3.7cm 0.cm 0.2cm},clip]{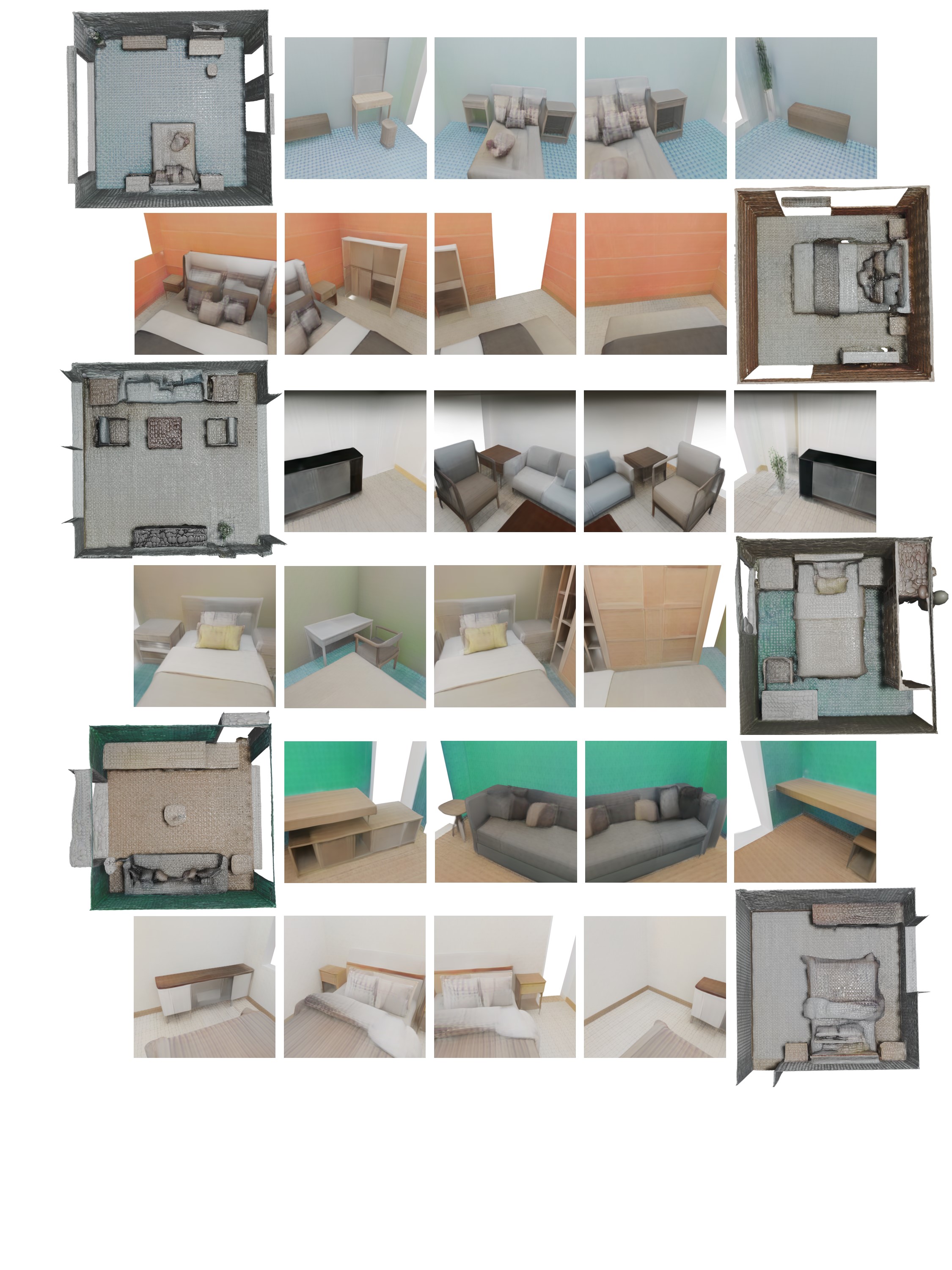} 
  \vspace{-0.3cm}
   \caption{Qualitative results on unconditional room generation on 3D-FRONT \cite{fu20213dfront}. Our method scales to room-size scenes and synthesizes plausible geometry and appearance. We visualize the generated 3D Gaussian ellipsoids and their renderings.
   }
   \label{fig:3dfront}
\end{figure*}
\begin{table}[tb]
\centering
\captionof{table}{Runtime comparison on ABO Tables~\cite{collins2022abo} dataset using one NVIDIA RTX A6000. By generating 3D Gaussians, our method enables much faster rendering speed. With a single forward pass, EG3D generates faster than diffusion-based approaches.}
\resizebox{\columnwidth}{!}{
\begin{tabular}{lccc}
\toprule
Method  & Generation time $\downarrow$ & \multicolumn{2}{c}{Rendering time $\downarrow$} \\ 
& per shape & \multicolumn{2}{c}{per frame} \\
\midrule
EG3D~\cite{Chan2022eg3d} & \textbf{6}ms & 23ms & @ $128\times128$ \\
DiffRF~\cite{muller2023diffrf}          & 21s & 48ms & @ $512\times512$ \\
Ours                                    & 13s & \textbf{0.91}ms & @ $512\times512$ \\
\bottomrule
\end{tabular}
}
\label{tab:runtime}
\end{table}
As shown in \cref{tab:photoshape,tab:abo}, our method leads to noticeable improvements compared to the baselines on all metrics. In particular, the perceptual metrics FID and KID show a large improvement, which indicates that our approach produces sharper, more detailed results.
This is confirmed by the qualitative comparisons in \cref{fig:photoshape_baseline,fig:abo_baseline}. All compared approaches produce plausible shapes. However, by generating 3D Gaussian primitives, our method is able to synthesize thinner structures, such as chair and table legs, where the DiffRF results are more coarse due to the limiting radiance field grid resolution. The GAN-based approach EG3D shows more artifacts and view-dependent inconsistencies, e.g., in the chair leg areas. 

\cref{tab:runtime} provides a runtime comparison. By synthesizing 3D Gaussians, which can be very efficiently rasterized, our method achieves significantly faster rendering speed, i.e., $\sim$50 times faster than DiffRF using radiance fields. The GAN-based EG3D generates shapes in a single network forward pass, hence has much faster generation time compared to diffusion-based approaches. Nonetheless, our method almost halves generation time compared to DiffRF. 

\subsection{Ablation Study}
To verify design choices of our method, we perform an ablation study on the ABO Tables dataset. The quantitative evaluation in \cref{tab:abo}, as well as the qualitative comparison in \cref{fig:abo_ablation} demonstrate that the full version of our method leads to the best performance. 

\begin{figure}[htb] % left bottom right top
  \centering
\includegraphics[width=\linewidth,trim={0cm 1.1cm 0cm 1.7cm},clip]{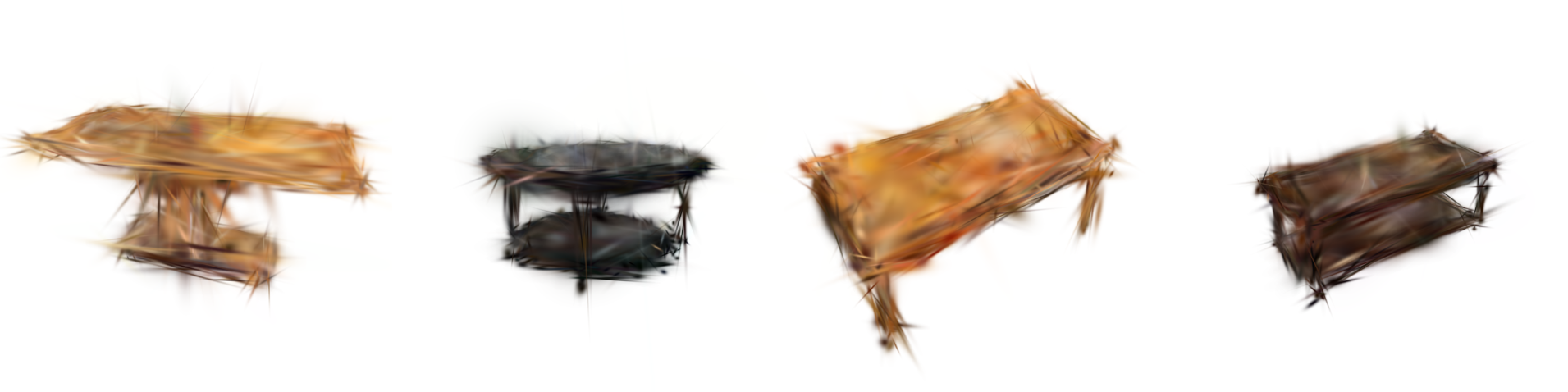}
   \caption{Ablation experiment ``w/o compression model'' struggles to generate coherent 3D Gaussians after the same training time as our complete method. }
   \label{fig:wo_vqvae}
\end{figure}

\paragraph{Without compression model} Omitting the 3D Gaussian compression model, i.e., training the diffusion model directly on optimized, grid-assigned 3D Gaussians (\cref{sssec:grid_aligned_3dg}) of low resolution ($32^3$), results in a drastic performance drop. We found that the diffusion model struggles to denoise this more complex, higher dimensional space of 3D Gaussian parameters $\boldsymbol{\theta}_{\kappa_i}$, compared to our vector-quantized latent features of 4 elements, where the codebook size allows for less than 10k alternative embeddings. After the same training time as our full method, the version ``w/o compression model'' still struggles to generate Gaussians that coherently describe the 3D shape (see examples in \cref{fig:wo_vqvae}). 

\paragraph{Without RGB} Dropping the RGB loss during training of the 3D Gaussian compression model leads to reduced perceptual and geometric metrics. Qualitatively, we observe that without the RGB loss the renderings tend to be more blurry and the color variety of the generated shapes seems reduced (\cref{fig:abo_ablation}). 

\paragraph{Without perceptual loss} The experiment without perceptual loss in the 3D Gaussian compression training shows a clear decrease in the performance measured by all metrics. The renderings lose sharpness, e.g., the wooden patterns on the tables in \cref{fig:abo_ablation} are no longer visible. 

\subsection{Other Qualitative Results}
\subsubsection{Unconditional Scene Generation} We showcase the ability of our latent 3D Gaussian diffusion to scale to room-size scenes. The sparse 3D Gaussians compression model enables flexible scaling to operate on scenes, which have $\sim$200k Gaussians, compared to $\sim$8k on objects, while the diffusion model can still operate on the same latent space dimension as for the object-level datasets. \cref{fig:3dfront} shows results on unconditional generation of rooms, where the model is trained on bedroom and living room style scenes from the 3D-FRONT~\cite{fu20213dfront} dataset. The generated scenes have plausible and varied configurations of furniture, and an accurate geometry, which is visible in the visualization of 3D Gaussian ellipsoids. 
\subsubsection{Nearest Neighbors in the Training Set}
\begin{figure}[htb] % left bottom right top
  \centering
\includegraphics[width=\linewidth,trim={0.6cm 1.2cm 17.cm 0.4cm},clip]{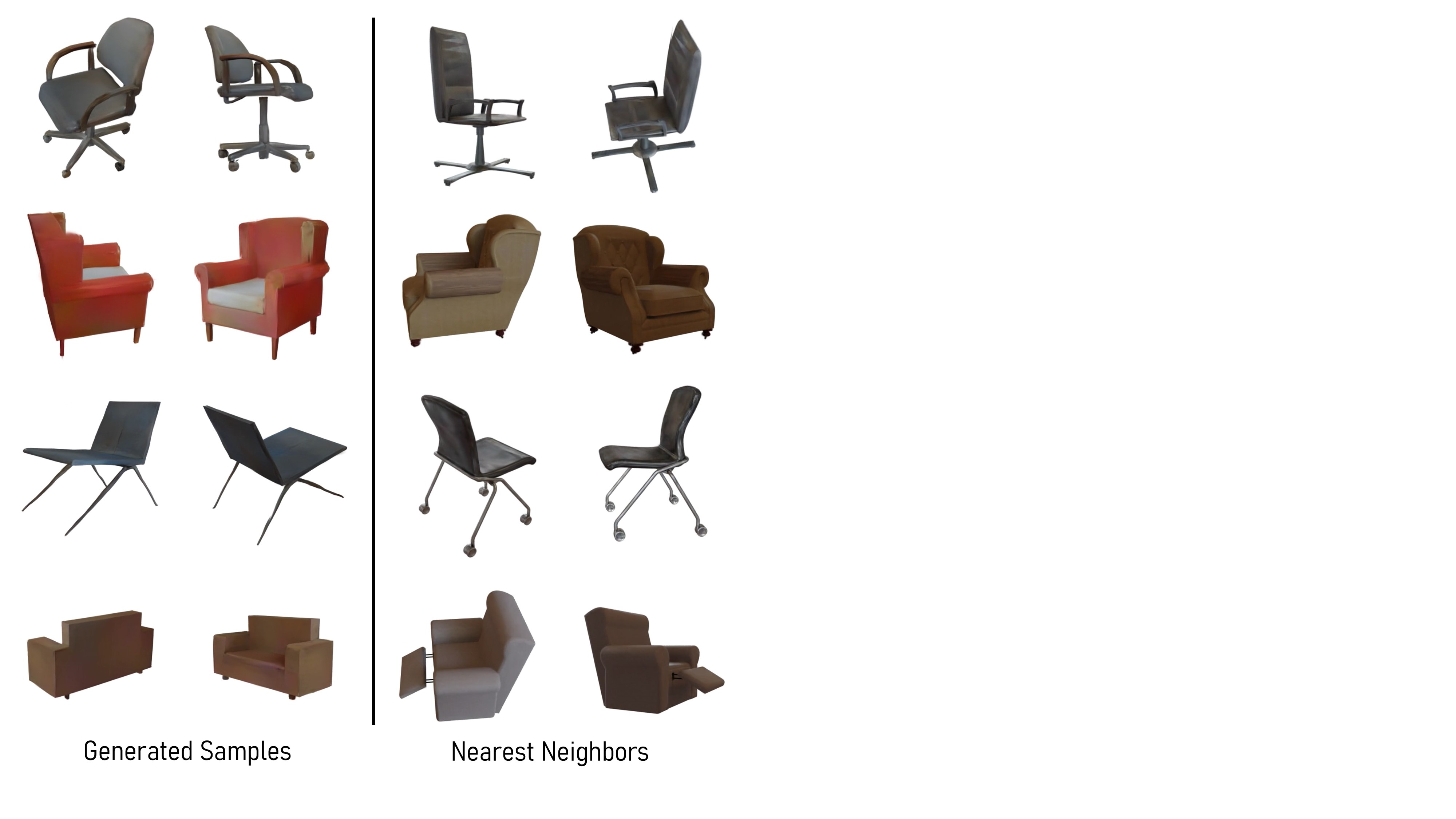}
   \caption{Visualization of geometric nearest neighbors in the training set using Chamfer Distance. Our approach can generate novel samples (left) that are different from their nearest neighbors in the training set (right).}
   \label{fig:nn}
\end{figure}

\cref{fig:nn} visualizes our generated chairs next to their nearest neighbors from the training set of optimized sparse grid-assigned 3D Gaussians. The geometric nearest neighbors are computed using Chamfer Distance on point clouds sampled from the 3D Gaussians. We observe that the generated chairs are substantially different from their nearest neighbors, indicating that the model does not purely retrieve from the training set, but generates novel shapes. 

\subsection{Limitations}
While our method is among the first to show its applicability to 3D scene generation at room-scale, we believe there are still significant open challenges.
One key ingredient towards achieving the scalability of our approach lies in the latent 3D scene representation of the 3D Gaussians. 
Here, analog to 2D image diffusion models~\cite{latent_diffusion}, larger neural network models will facilitate the creation of outputs of larger scene extents and higher visual fidelity.
In this context, available computational resources was a major bottleneck that limited further exploration. 
However, at the same time, we believe that additional training strategies, e.g., exploiting spatial subdivision strategies of 3D spaces, could further alleviate memory and computational limitations.

At the same time, our method is currently trained on synthetic datasets such as PhotoShape~\cite{photoshape2018}, ABO~\cite{collins2022abo}, or 3D-FRONT~\cite{fu20213dfront}. 
Here, we can see a future potential on real-world datasets that provide ground truth 3D supervision at the scene level. 
Unfortunately, 3D datasets with high-fidelity DSLR captures (which is required to reconstruct the Gaussian ground truth pairs), such as Tanks and Temples~\cite{knapitsch2017tanks} or ScanNet++~\cite{yeshwanthliu2023scannetpp} are still relatively limited in terms of the number of available 3D scenes.
\section{Conclusion}

We have presented \OURS, a novel generative approach that models a 3D scene distribution represented by 3D Gaussians.
The core idea of our method is a latent 3D diffusion model whose latent space is learned by a VQ-VAE for which we propose a sparse convolutional 3D architecture.
This facilitates the scalability of our method and significantly improves the visual quality over existing works. 
For instance, in comparison to NeRF-based generators, such as DiffRF~\cite{muller2023diffrf}, L3DG can be rendered faster and thus trained on larger scenes.
In particular, this allows us to showcase a first step towards room-scale scene generation.
Overall, we believe that our method is an important stepping stone to support the 3D content generation process along a wide range of applications in computer graphics. 
\begin{acks}
This work was funded by a Meta SRA. Matthias Nießner was also supported by the ERC Starting Grant Scan2CAD (804724) and Angela Dai was supported by the ERC Starting Grant SpatialSem (101076253).
\end{acks}
\bibliographystyle{ACM-Reference-Format}
\bibliography{main}

\end{document}